\documentclass{article}
\usepackage[square,sort,comma,numbers]{natbib}
\usepackage[preprint]{neurips_2023}

\usepackage[utf8]{inputenc} % allow utf-8 input
\usepackage[T1]{fontenc}    % use 8-bit T1 fonts
\usepackage{hyperref}       % hyperlinks
\usepackage{url}            % simple URL typesetting
\usepackage{booktabs}       % professional-quality tables
\usepackage{multirow} % For multi-row cells
\usepackage{multicol} % For multi-column cells
\usepackage{pifont}   % For checkmarks and cross marks
\usepackage{adjustbox}
\usepackage{amsfonts}       % blackboard math symbols
\usepackage{nicefrac}       % compact symbols for 1/2, etc.
\usepackage{microtype}      % microtypography
\usepackage{xcolor}         % colors
\usepackage{subfiles} 
\usepackage{cite}
\usepackage{wrapfig}
\usepackage{upgreek}
\usepackage{bm}
\usepackage{mathtools}
\DeclareMathOperator*{\argmax}{arg\,max}
\DeclareMathOperator*{\argmin}{arg\,min}
\newcommand{\vect}[1]{\boldsymbol{#1}}
\usepackage{algorithm,algpseudocode}
\usepackage{algorithmicx}

\title{Sparse but Strong: Crafting Adversarially Robust Graph Lottery Tickets}

\author{%
  Subhajit Dutta Chowdhury$^1$\thanks{Equal Contribution} , Zhiyu Ni$^{1\ast}$, Qingyuan Peng$^{1\ast}$, Souvik Kundu$^2$, Pierluigi Nuzzo$^1$ \\
  $^1$ University of Southern California, $^2$ Intel Labs\\
}

\begin{document}

\maketitle

\begin{abstract}

Graph Lottery Tickets (GLTs), comprising a sparse adjacency matrix and a sparse graph neural network (GNN), can significantly reduce the inference latency and compute footprint compared to their dense counterparts. Despite these benefits, their performance against adversarial structure perturbations remains to be fully explored. In this work, we first investigate the resilience of GLTs against different structure perturbation attacks and observe that they are highly vulnerable and show a large drop in classification accuracy. Based on this observation, we then present an adversarially robust graph sparsification (ARGS) framework that prunes the adjacency matrix and the GNN weights by optimizing a novel loss function capturing the graph homophily property and information associated with both the true labels of the train nodes and the pseudo labels of the test nodes. By iteratively applying ARGS to prune both the perturbed graph adjacency matrix and the GNN model weights, we can find adversarially robust graph lottery tickets that are highly sparse yet achieve competitive performance under different untargeted training-time structure attacks. Evaluations conducted on various benchmarks, considering different poisoning structure attacks, namely, PGD, MetaAttack, Meta-PGD, and PR-BCD demonstrate that the GLTs generated by ARGS can significantly improve the robustness, even when subjected to high levels of sparsity. 
\end{abstract}

\section{Introduction}

Graph neural networks (GNNs)~\citep{hamilton2017inductive,kipf2016semi,velivckovic2017graph,zhou2020graph, zhang2020deep} achieve state-of-the-art 
performance on various graph-based tasks like semi-supervised node classification~\citep{kipf2016semi,hamilton2017inductive,velivckovic2017graph}, link prediction~\citep{zhang2018link}, and graph classification~\citep{ying2018hierarchical}. The success of GNNs is attributed to the neural message-passing scheme in which each node updates its feature by recursively aggregating and transforming the features of its neighbors. However, the effectiveness of GNNs, when scaled up to large and densely connected graphs, is adversely affected by the high training cost, high inference latency, and substantial memory consumption. Unified graph sparsification (UGS)~\citep{chen2021unified} addresses this concern by simultaneously pruning the input graph adjacency matrix and the GNN to 
% show the existence of 
achieve a graph lottery ticket (GLT), a pair of sparse graph adjacency matrix and GNN model, which can potentially accelerate inference without compromising model performance.

Recent studies reveal that GNNs are vulnerable to adversarial attacks~\citep{dai2018adversarial, wu2019adversarial, zugner2020adversarial,mujkanovic2022defenses, jin2020adversarial}. An adversarial attack introduces unnoticeable perturbations to the graph structure or node features. These perturbations increase the distribution shift between train nodes and test nodes, fooling the GNN to misclassify nodes in the graph~\citep{lirevisiting}. When compared to node features, altering the graph structure has a more significant impact on the classification accuracy. To counter these attacks, many defense techniques have been developed that try to improve the classification accuracy of GNNs either by cleaning the perturbed graph structure~\citep{zhu2021deep, zhang2020gnnguard,wu2019adversarial, jin2020graph} or by introducing new learning approaches~\citep{lirevisiting,feng2020graph}. On the other hand, while GLTs demonstrate strong performance on original benign graph data, their performance in the presence of adversarial structure perturbations remains largely unexplored. \emph{Finding adversarially robust GLTs is key to enabling efficient GNN inference under adversarial threats.}

\begin{wrapfigure}{r}{0.47\textwidth}
  \begin{center}
  \vspace{-5mm}
    \includegraphics[width=0.47\textwidth]{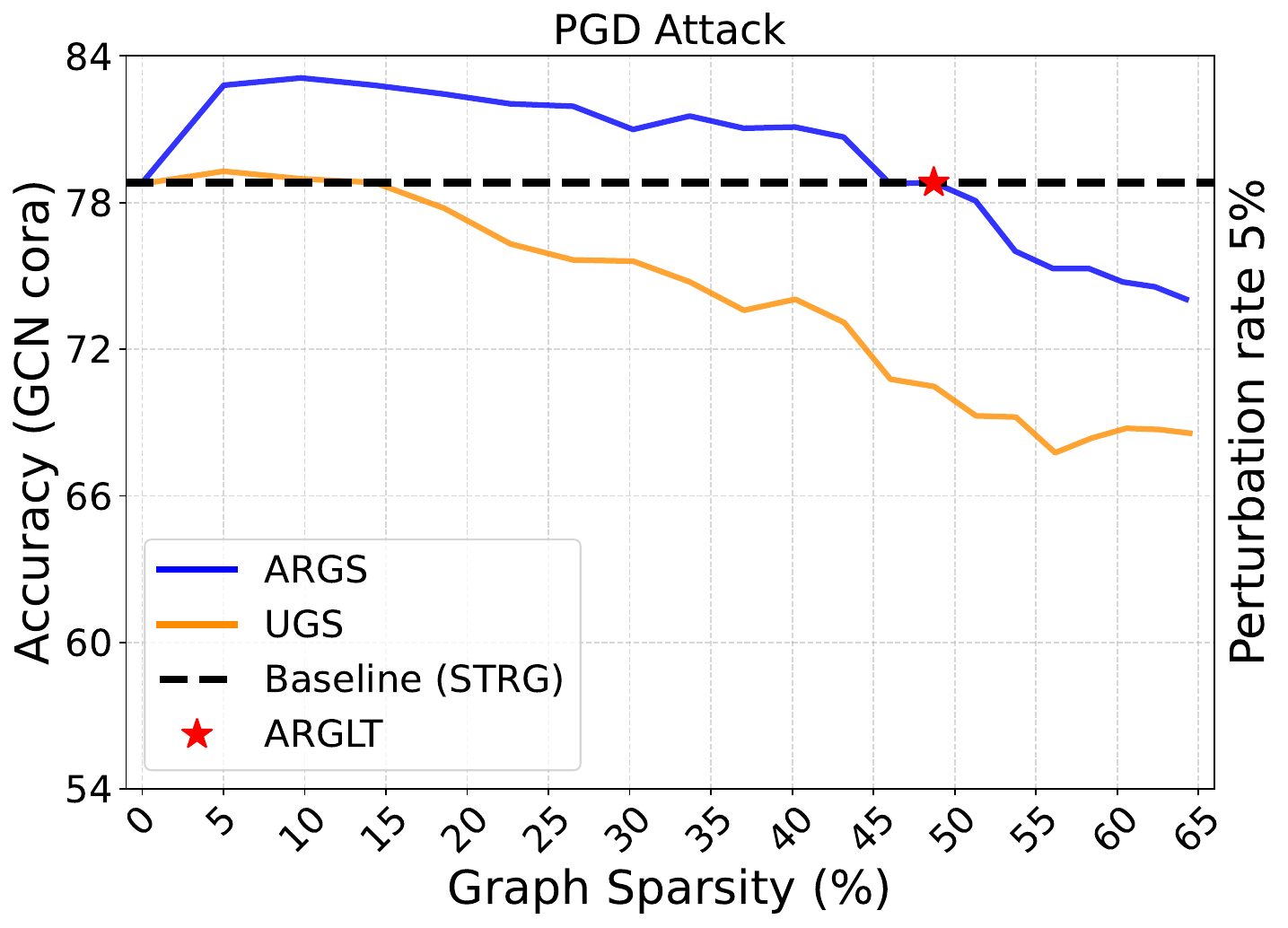}
  \end{center}
  \caption{Comparison of different graph sparsification techniques in accuracy vs. graph sparsity trade-off plot. ARGS archives similar accuracy with $30\%$ more sparsity compared to UGS for the Cora dataset under PGD attack.}
  \label{fig:intro_figure}
\end{wrapfigure}

To this end, we first empirically investigate the resilience of GLTs identified by UGS against different structure perturbation attacks~\citep{zugner2020adversarial,liu2019unified,mujkanovic2022defenses}. Then, we present ARGS, adversarially robust graph sparsification, an end-to-end optimization framework that, given an adversarially perturbed graph, iteratively prunes the graph adjacency matrix and the GNN model weights to generate an adversarially robust graph lottery ticket (ARGLT) that achieves competitive classification accuracy while exhibiting high levels of sparsity. In this work, we consider a widely adopted setting for adversarial attacks, i.e., poisoning adversarial attacks on the graph structure. Attacks like the projected gradient descent (PGD) topology attack~\citep{wu2019adversarial}, the meta-learning-based graph attack (MetaAttack)~\citep{zugner2020adversarial}, and Meta-PGD 
% that applies PGD on the meta-gradients~
\citep{mujkanovic2022defenses} often introduce many of the edge modifications around the training nodes~\citep{lirevisiting} while the local structure of the test nodes is less affected. Moreover, the adversarial edges introduced are often between nodes with dissimilar features~\citep{wu2019adversarial}.
We leverage this information to formulate a new loss function that better guides the pruning of the adversarial edges in the graph and weights of the GNN. Additionally, we use self-learning to train pruned GNNs on sparse graph structures, which improves the classification accuracy of the GLTs. 

Our proposal is experimentally verified across various GNN architectures on different graph datasets (Cora, Citeseer, PubMed, and OGBN-ArXiv) attacked by poisoning attacks (PGD, MetaAttack, Meta-PGD, PR-BCD \citep{geisler2021robustness}) for the node classification task. Results show that ARGS is widely applicable for sparsifying the GNN and the graph adjacency matrix when the graph structure has been adversarially attacked. By iteratively applying ARGS, ARGLTs can be broadly located across the 4 graph datasets with substantially reduced inference costs and unimpaired performance. The ARGLTs achieve $23\%-61\%$ sparsity on graphs and $64\% - 98\%$ sparsity on GNN models, at little to no adversarial performance degradation. Figure~\ref{fig:intro_figure} shows that, for node classification on Cora attacked by the PGD attack (5\% perturbation), our ARGLT achieves similar accuracy to that of the full model and graph even with high graph and model sparsity of $49\%$ and $95\%$, respectively. When compared to the GLTs identified by UGS, the ARGLTs on average achieve the same accuracy with $\sim 2.4\times$ more graph sparsity and $\sim 2.3\times$ more model sparsity.

%If needed:
%In summary, our contributions are:
%\begin{itemize}
%  \item L
%  \item
%  \item
%\end{itemize}

\section{Related Work}

\textbf{Graph Lottery Ticket Hypothesis.} The lottery ticket hypothesis (LTH)~\citep{frankle2018lottery} conjectures that there exist small sub-networks, dubbed as lottery tickets (LTs), within a dense randomly initialized network, that can be trained in isolation to achieve comparable accuracy to their dense counterparts. UGS made it possible to extend the LTH to GNNs~\citep{chen2021unified}, showing the existence of GLTs that can accelerate GNN inference. A GNN sub-network along with a sparse graph is defined as a GLT if the sub-network with the original initialization trained on the sparsified graph has a matching test accuracy to the original unpruned GNN trained on the full graph. Specifically, during end-to-end training, UGS applies two differentiable binary mask tensors to the graph adjacency matrix and the GNN model weights, respectively. After training, the lowest-magnitude elements are removed and the corresponding mask location is updated to 0, eliminating the low-scored edges and weights from the adjacency matrix and the GNN, respectively. The sparse GNN weight parameters are then rewound to their original initialization. To identify the GLTs, the UGS algorithm is applied in an iterative fashion until pre-defined graph and weight sparsity levels are reached. Experimental results show that UGS can significantly trim down the inference computational cost without compromising the predictive accuracy. In this work, we aim to find GLTs for datasets that have been adversarially perturbed. When we apply the UGS algorithm directly to the perturbed graphs, the performance accuracy of the GLTs is substantially low compared to their clean counterparts, calling for new optimization method to find adversarially robust GLTs.

% \textbf{Adversarial Attacks on Graphs.} Based on the time of the attack, adversarial attacks on graphs can be classified as poisoning attacks, which perturb the graph at train time, and evasion attacks, perturbing the graph at test time. Poisoning attacks can be further classified into targeted attacks and global attacks~\citep{liu2019unified}. A targeted attack deceives the model to misclassify a specific node (i.e., target node)~\citep{zugner2018adversarial} while a global attack degrades the overall performance of the trained model~\citep{zugner2020adversarial,wu2019adversarial}. In the case of graph data, an attacker can modify the node features, the discrete graph structure, or both. Different attacks show that structure perturbation is more effective compared to modifying the node features. Examples of global poisoning attacks include the MetaAttack~\citep{zugner2020adversarial}, PGD attack~\citep{wu2019adversarial}, and Meta-PGD attack\citep{mujkanovic2022defenses}. All these gradient-based attacks treat the adjacency matrix as a parameter tensor and modify it via scaled gradient-based perturbations that aim to maximize the loss, thus resulting in degradation of the GNN prediction accuracy. This objective can be formulated as a bi-level optimization problem.
% In this work, we consider global graph structure poisoning attacks and develop ARGS for finding RGLTs since these attacks have shown remarkable effectiveness in reducing the classification accuracy of different GNNs compared to other forms of attacks.

\textbf{Adversarial Attacks on Graphs.} Adversarial attacks on graphs can be classified as poisoning attacks, perturbing the graph at train time, and evasion attacks, perturbing the graph at test time. Both poisoning and evasion attacks can be targeted or global attacks~\citep{liu2019unified}. A targeted attack deceives the model to misclassify a specific node~\citep{zugner2018adversarial, bojchevski2019adversarial}. A global attack degrades the overall performance of the model~\citep{zugner2020adversarial,wu2019adversarial}. Depending on the amount of information available, the existing attacks can be categorized into white-box attacks, practical black-box attacks, and restricted black-box attacks~\citep{zugner2018adversarial,chang2020restricted}. An attacker can modify the node features, the discrete graph structure, or both. Different attacks show that structure perturbation is more effective when compared to modifying the node features. Examples of global poisoning attacks include the MetaAttack~\citep{zugner2020adversarial}, PGD attack~\citep{wu2019adversarial}, and PR-BCD attack~\citep{geisler2021robustness}. Gradient-based attacks like PGD and MetaAttack treat the adjacency matrix as a parameter tensor and modify it via scaled gradient-based perturbations that aim to maximize the loss, thus resulting in degradation of the GNN prediction accuracy. 
% However, these attacks do not scale. 
PR-BCD~\citep{geisler2021robustness} is a more scalable first-order optimization attack that can scale up to large datasets like OGBN-ArXiv~\citep{hu2020open}. % As evident in the literature, 
Global poisoning attacks are highly effective in reducing the classification accuracy of different GNNs and are typically more challenging to counter since they modify the graph structure before training~\citep{zhu2021improving}. Therefore, we consider global graph structure poisoning attacks.

\textbf{Defenses on Graphs.} Several approaches for improving the robustness of GNNs have been developed to combat adversarial attacks on graphs~\citep{tang2020transferring, entezari2020all, zhu2019robust, jin2020graph, zhang2020gnnguard, wu2019adversarial}. Many of these techniques try to improve the classification accuracy by preprocessing the graph structure,  i.e., they detect the potential adversarial edges and assign these edges lower weights or even remove them. Jaccard-GCN~\citep{wu2019adversarial} removes all edges between nodes whose features exhibit a Jaccard similarity below a certain threshold by leveraging the homophily property of graphs. SVD-GCN~\citep{entezari2020all} replaces the adjacency matrix with a low-rank approximation since many real-world clean graphs are low-rank and attacks tend to disproportionately affect the high-frequency spectrum of the adjacency matrix. ProGNN~\citep{jin2020graph} leverages low-rank, sparsity, and feature smoothness properties of graphs to clean the perturbed adjacency matrix and improve the trained GNN performance. GNNGuard~\citep{zhang2020gnnguard} learns weights for the edges in each message passing aggregation via cosine-similarity and penalizes the adversarial edges by either filtering them or assigning less weight. Other techniques try to improve the GNN performance by enhancing model training through data augmentation~\citep{li2022reliable, feng2020graph}, adversarial training~\citep{wu2019adversarial}, self-learning~\citep{lirevisiting}, or by developing novel GNN layers, e.g., RGCN~\citep{zhu2019robust}, which adopts Gaussian distributions as the hidden representations of the nodes in each convolutional layer to absorb the effect of an attack. Graph preprocessing tends to remove only a small fraction of edges from the adjacency matrix. Thus, the existing defense techniques often lead to high inference latency and are not scalable to real-world graphs. Differently from the existing defense methods that are built on dense GNNs with dense or nearly dense adjacency matrices,  we aim to improve the robustness of low-latency sparse GNNs with highly sparse adjacency matrices. However, as robustness generally requires more non-zero parameters, yielding sufficiently sparse robust GLTs remains a challenge.

\section{Methodology}
\textbf{Notations.} Let $ \mathcal{G = \{V,E\}} $ represent an undirected graph with $ \mathcal{|V|} $ nodes and $ \mathcal{|E|} $ edges.
The topology of the graph can be represented with an adjacency matrix $\vect{A} \in \mathbb{R}^{\mathcal{|V| \times |V|}}$, where $ \vect{A}_{ij} = 1 $ if there is an edge $ e_{i,j} \in \mathcal{E}$ between nodes $v_{i}$ and $v_{j}$, while $\vect{A}_{ij} = 0 $ otherwise. 
% if the nodes are not connected. 
Each node $v_{i} \in \mathcal{V}$ has a feature vector $\textbf{x}_{i} \in \mathbb{R}^F$, where $F$ is the number of node features. Let $ \vect{X} \in \mathbb{R}^{\mathcal{|V|}\times F} $ and $\vect{Y} \in \mathbb{R}^{\mathcal{|V|} \times C} $ denote the feature matrix and the labels of all nodes in the graph, respectively, where $ C $ is the total number of classes. In this paper, we will also represent a graph as a pair $ \{\vect{A, X}\} $. In the case of message-passing GNN, the representation of a node $ v_i$ is iteratively updated by aggregating and transforming the representations of its neighbors. As an example, a two-layer\citep{kipf2016semi} GNN can be specified as
\begin{equation}
    \vect{Z} = f(\{\vect{A}, \vect{X}\}, \bm{\Theta}) = \mathcal{S}(\hat{\vect{A}}\sigma(\vect{\hat{A}}\vect{X}\vect{W}_{(0)})\vect{W}_{(1)}),
    \label{eq:1}
\end{equation}
where $ \vect{Z}$ is the prediction,  
% $f(\cdot, \bm{\Theta)}$, 
%$ f(\mathcal{G},\bm{\Theta}) $. 
$ \bm{\Theta} = (\vect{W}_{0}, \vect{W}_1) $ are the weights, $ \sigma(.)$ is the activation function, e.g., a rectified linear unit (ReLU), $\mathcal{S}(.)$ is the softmax function, $\vect{\hat{A}} = \vect{\Tilde{D}^{-\frac{1}{2}}}(\vect{A}+\vect{I})\vect{\Tilde{D}^{-\frac{1}{2}}}$ is the normalized adjacency matrix with self-loops, and $\vect{\Tilde{D}}$ is the degree matrix of $\vect{A+I}$. We consider the transductive semi-supervised node classification (SSNC) task for which the cross-entropy (CE) loss over labeled nodes is given by
\begin{equation}
    \mathcal{L}_{0}(f(\{\vect{A}, \vect{X}\}, \bm{\Theta})) = - \sum_{l \in \mathcal{Y}_{TL}} \sum_{j=1}^{C} \vect{Y}_{l_{j}}\log ( \vect{Z}_{l_{j}} ) ,
    \label{eq:2}
\end{equation}
where $ \mathcal{Y}_{TL}$ is the set of train node indices, $ C $ is the total number of classes, and $ \vect{Y}_{l} $ is the one-hot encoded label of node $ v_{l} $. 

\textbf{Graph Lottery Tickets}. A GLT consists of a sparsified graph, obtained by pruning some edges in $\mathcal{G}$, and a GNN sub-network, with the original initialization, that can be retrained to achieve comparable performance to the original GNN trained on the full graph, where performance is measured in terms of test accuracy. Given a GNN $ f(\cdot, \bm{\Theta)}$ and a graph $\mathcal{G} = \{\vect{A, X}\} $, the associated GNN sub-network and the sparsified graph can be represented as $ f(\cdot, \vect{m}_{\theta} \odot \bm{\Theta)}$ and $ \mathcal{G}_{s} = \{ \vect{m}_{g} \odot \vect{A, X}\} $, respectively, where $\vect{m}_g$ and $\vect{m}_{\theta}$ are differentiable masks applied to the adjacency matrix $\vect{A}$ and the model weights $\bm{\Theta}$, respectively, and $\odot$ is the element-wise product. UGS\citep{chen2021unified} finds the two masks $\vect{m}_g$ and $\vect{m}_{\theta}$ such that the sub-network $ f(\cdot, \vect{m}_{\theta} \odot \bm{\Theta)}$ along with the sparsified graph $ \mathcal{G}_{s}$ can be trained to a similar accuracy as $ f(., \bm{\Theta)}$.

\textbf{Poisoning Attack on Graphs.}  In this work, we investigate the robustness of GLTs under non-targeted poisoning attacks modifying the structure of the graph. In the case of a poisoning attack, GNNs are trained on a graph that attackers maliciously modify. The aim of the attacker is to find an optimal perturbed $\vect{A^{'}}$ that fools the GNN into making incorrect predictions. This can be formulated as a bi-level optimization problem~\citep{zugner2018adversarial,zugner2020adversarial}:
\begin{equation}
    \begin{gathered}
     \argmax_{\vect{A}^{'} \in \Phi(\vect{A})} \mathcal{L}_{atk}(f(\{\vect{A}^{'}, \vect{X}\}, \bm{\Theta}^*)) \\
     \mathrm{s.t.} \;\;\;\; \bm{\Theta}^{*} = \argmin_{\bm{\Theta}} \mathcal{L}_{0}(f(\{\vect{A}^{'}, \vect{X}\}, \bm{\Theta}))
    \end{gathered}
    \label{eq:3}
\end{equation}

where $\Phi(\vect{A})$ is the set of adjacency matrices that fit the constraint $\frac{{||\vect{A}^{'} - \vect{A}||}_{0}}{||\vect{A}||_{0}} \leq \Delta $, $\mathcal{L}_{atk}$ is the attack loss function, $\Delta$ is the perturbation rate, and $\bm{\Theta}^{*}$ is the optimal parameter for the GNN on the perturbed graph.

\begin{figure}[t]
  \centering
 \includegraphics[width=1\columnwidth]{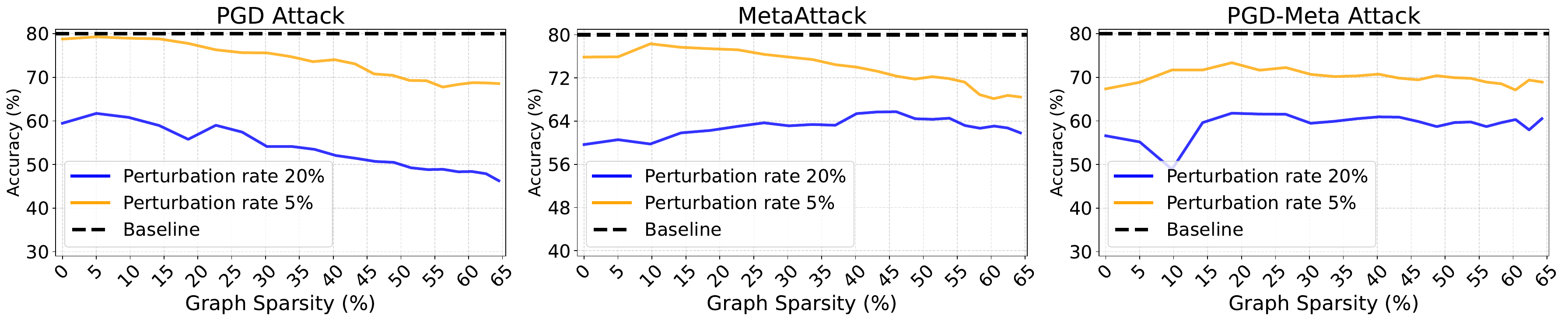} 
  \caption{Classification accuracy of GLTs generated using UGS for the Cora dataset under PGD (left),  MetaAttack (middle), and Meta-PGD attack (right) with 5\% and 20\% perturbation rates.
  The baseline refers to the accuracy on the clean graph.}
  \label{fig:ugs_eval}
 \end{figure}

\subsection{UGS Analysis Under Adversarial Attacks} 

We perform the MetaAttack~\citep{zugner2020adversarial}, PGD~\citep{wu2019adversarial}, and Meta-PGD~\citep{mujkanovic2022defenses} attacks on the Cora dataset with different perturbation rates. Then, we apply UGS on these perturbed graphs to find the GLTs. As evident from Fig.~\ref{fig:ugs_eval}, the classification accuracy of the GLTs identified by UGS is lower than the clean graph accuracy. The difference increases substantially when the perturbation rate increases. For example, in the PGD attack,  when the graph sparsity is $30\%$, at $5\%$ perturbation, the accuracy drop is $6\%$. This drop increases to $25\%$ when the perturbation rate is $20\%.$ Moreover, for $20\%$ perturbation rate, even with $0\%$ sparsity, the accuracy of the GNN is around 20\% lower than that of the clean graph accuracy. While  UGS removes edges from the perturbed adjacency matrix, 
% it is evident from Fig.~\ref{fig:ugs_eval} that 
it may not effectively remove the adversarially perturbed edges. A na\"{i}ve application of UGS may not be sufficient to improve the adversarial robustness of the GLTs. Consequently, there is a need for an adversarially robust UGS technique that can efficiently remove the edges affected by  adversarial perturbations while pruning the adjacency matrix and the associated GNN, along with improved adversarial training, allowing the dual benefits of improved robustness and inference latency.

\subsection{Analyzing the Impact of Adversarial Attacks on the Graph Properties}

\begin{wrapfigure}{r}{0.44\textwidth}
  \begin{center}
  \vspace{-10mm}
    \includegraphics[width=0.44\textwidth]{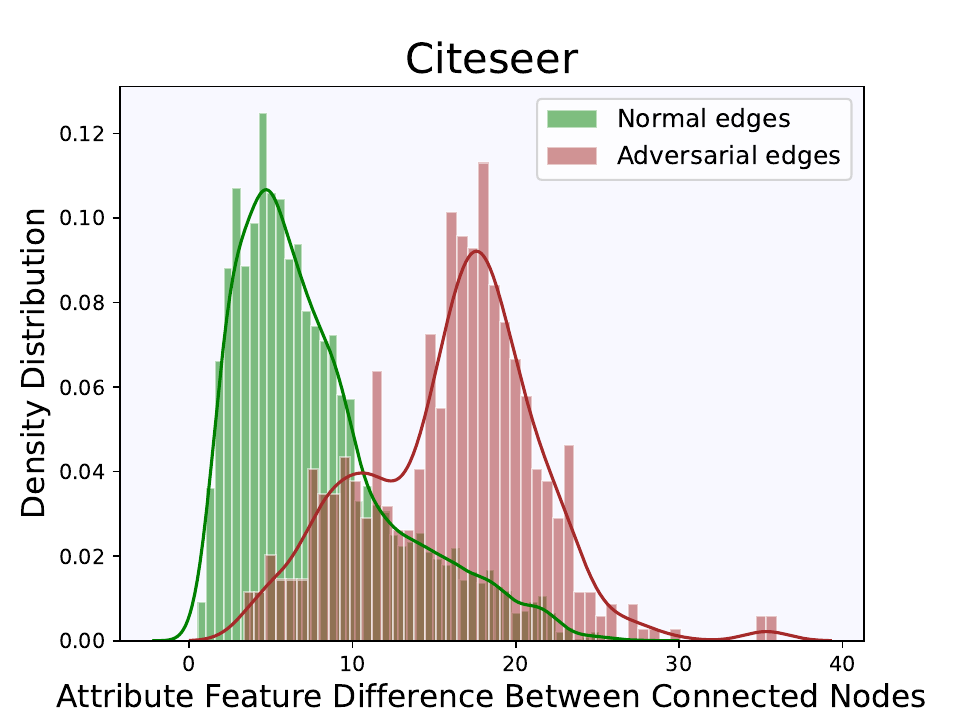}
  \end{center}
  \vspace{-5pt}
  \caption{Density distribution of attribute feature differences between connected nodes in perturbed Citeseer graph dataset.}
  \label{fig:feature_diff}
\end{wrapfigure}

Adversarial attacks like MetaAttack, PGD, and PR-BCD poison the graph structure by either introducing new edges or deleting existing edges, resulting in changes in the original graph properties. We analyze the difference in the attribute features of the nodes connected by the clean and adversarial edges. Figure~\ref{fig:feature_diff} depicts the density distribution of the attribute feature difference between connected nodes in the Citeseer graph dataset attacked by the PGD attack. We can observe from Figure~\ref{fig:feature_diff} that the attack tends to connect nodes with large attribute feature differences. A defense technique can potentially leverage this information to differentiate between the benign and adversarial edges in the graph. ARGS uses this observation to iteratively prune the adversarial edges from homophilic graphs.

\subsection{Adversarially Robust Graph Sparsification}

{
We present ARGS, a sparsification technique that simultaneously reduces edges in
$\mathcal{G}$ and GNN parameters in $\bm{\Theta}$ under adversarial attack conditions to effectively accelerate GNN inference yet maintain robust classification accuracy. ARGS reformulates the loss function to include (a) a CE loss term on the train nodes, (b) a CE loss term on a set of test nodes, and (c) a square loss term on all edges. Pruning the edges based on this combined loss function results in the removal of adversarial as well as less-important non-adversarial edges from the graph.

% \setlength{\abovecaptionskip}{3pt}
% \setlength{\belowcaptionskip}{3pt}
% \begin{wrapfigure}{r}{0.45\textwidth}
%   \begin{center}
%      \includegraphics[width=0.45\textwidth]{figures/edge_distribution.pdf}
%   \end{center}
%   \caption{Evolution of adversarial edges in Cora dataset (attacked by PGD, $20\%$ perturbation) as we apply ARGS to prune the graph. Train-Train edges connect two nodes from the train set. Train-Test edges connect two nodes from the train and test set, respectively. Test-Test edges connect two nodes from the test set.}
%   \label{fig:edge_distribtuion}
%   \vspace{-4mm}
% \end{wrapfigure}

\textit{Removing Edges Around Train Nodes.} Poisoning attacks like the MetaAttack and the PGD attack tend to modify more the local structure around the train nodes than that around the test nodes~\citep{lirevisiting}. Specifically, a large portion of the modifications is introduced to the edges connecting a train node to a test node or a train node to another train node. We include a CE loss term associated with the train nodes, as defined in~\eqref{eq:2} in our objective function to account for the edges surrounding the train nodes. These edges include both adversarial and non-adversarial edges.

\textit{Removing Adversarial Edges.} Adversarial attacks on graphs tend to connect nodes with distinct features~\citep{wu2019adversarial}. However, it has been observed in numerous application domains involving social graphs, web page graphs, and citation graphs, that connected nodes within a graph tend to exhibit similar features~\citep{li2022reliable,mcpherson2001birds, kipf2016semi}. To help remove the adversarial edges and encourage feature smoothness, we include the following loss to our objective function:
\begin{equation}
    \mathcal{L}_{fs}(\vect{A}^{'}, \vect{X}) = \frac{1}{2} \sum_{i,j = 1} \vect{A}^{'}_{ij} (\vect{x_i} -\vect{x_j})^{2} ,
    \label{eq:4}
\end{equation}
where $\vect{A}^{'}$ is the perturbed adjacency matrix and $(\vect{x}_i - \vect{x}_j)^{2}$ measures the feature difference between $\vect{x}_i$ and $\vect{x}_j$. Minimizing this loss function encourages removing edges that connect dissimilar nodes since they contribute a higher loss term.

\textit{Removing Edges Around Test Nodes.} Removal of edges tends to be random in later iterations of UGS~\citep{hui2023rethinking} since only a fraction of edges in $\mathcal{G}$ is related to the train nodes and directly impacts the corresponding CE loss. To better guide the edge removal around the test nodes, we also introduce a CE loss term for these nodes. However, the labels of the test nodes are unknown. We can then leverage the fact that structure poisoning attacks modify only the structure surrounding the train nodes, while their features and labels remain clean. Therefore, we first train a 
simple multi-layer perceptron (MLP) with 2 layers on the train nodes. MLPs only use the node features for training. We then use the trained MLP to predict the labels for the test nodes. We call these labels pseudo labels. Finally, we use the test nodes for which the MLP has high prediction confidence for computing the test node CE loss term.  
Let $\mathcal{Y}_{PL}$ be the set of test nodes for which the MLP prediction confidence is high and $\vect{Y}_{mlp}$ be the prediction by the MLP. The CE loss is given by
\begin{equation}
    \mathcal{L}_{1}(f(\{\vect{A}^{'}, \vect{X}\}, \bm{\Theta})) = - \sum_{l \in \mathcal{Y}_{TL}} \sum_{j=1}^{C} \vect{Y}_{mlp_{l_{j}}}\log ( \vect{Z}_{l_{j}} ).
    \label{eq:5}
\end{equation}

%We leverage self-learning to appropriately guide the removal of edges not related to the train nodes. We first train a simple 2-layer multi-layer perceptron (MLP) with the train nodes. In fact, the structure poisoning attack modifies only the structure surrounding the train nodes, while their features and labels remain clean. Then, we use the trained MLP to get the pseudo labels of the test nodes. We include the CE loss associated with the test nodes for which the MLP has high prediction confidence in our objective function to better control the edge removal from the graph. Let $\mathcal{Y}_{PL}$ be the set of test nodes for which the MLP prediction confidence is high and $\vect{Y}_{mlp}$ be the prediction by the MLP. Then  the CE loss is given by
In summary, the complete loss function that ARGS optimizes is
\begin{equation}
    \begin{gathered}
        \mathcal{L}_{ARGS} = \alpha \mathcal{L}_{0}(f(\{\vect{m}_{g}\odot\vect{A}^{'}, \vect{X}\}, \vect{m}_{\theta}\odot\bm{\Theta})) + \beta \mathcal{L}_{fs}(\vect{m}_{g}\odot\vect{A}^{'}, \vect{X}) \\
        +\gamma \mathcal{L}_{1}(f(\{\vect{m}_{g}\odot\vect{A}^{'}, \vect{X}\}, \vect{m}_{\theta}\odot\bm{\Theta})) + \lambda_{1}||\vect{m}_{g}||_{1} + \lambda_{2}||\vect{m}_{\theta}||_{1},
    \end{gathered}
    \label{eq:6}
\end{equation}
where $\beta, \gamma, \lambda_{1}$, and $\lambda_2$ are the hyperparameters. The value of $\alpha, \gamma$ is 1. $\lambda_{1}$ and $\lambda_{2}$ are the $l_1$ regularizers for $\vect{m}_{g}, \vect{m}_{\theta}$, respectively. After the training is complete, $p_{g}$, the lowest percentage  of elements of $\vect{m}_{g}$, and $p_{\theta}$, the lowest percentage of elements of $\vect{m}_{\theta}$, are set to $0$. Then, the updated masks are applied to prune $\vect{A}$ and $\bm{\Theta}$, and the weights of the GNN are rewound to their original initialization value to generate the ARGLT.
%\souvik{After each training iteration, the lowest $p_{g}\%$ elements of $\vect{m}_{g}$, and $p_{\theta}\%$ elements of $\vect{m}_{\theta}$ are made zero, that gets added to the zero elements aggregated over previous iterations to iteratively reach a certain sparsity ratio.}
%The binary mask $\vect{m}_{\theta}$ also gets updated to be sparser by $p_{\theta}$ fraction where indices corresponding to the lowest $p_{\theta}$ fraction of non-zero magnitude weights are made zero.} 
%Then, the two sparse masks are applied to prune $\vect{A}$ and $\bm{\Theta}$ and $\bm{\Theta}$ is rewound to original initialization. 

\setlength{\abovecaptionskip}{3pt}
\setlength{\belowcaptionskip}{3pt}
\begin{wrapfigure}{r}{0.45\textwidth}
    \vspace{-5mm}
  \begin{center}
     \includegraphics[width=0.45\textwidth]{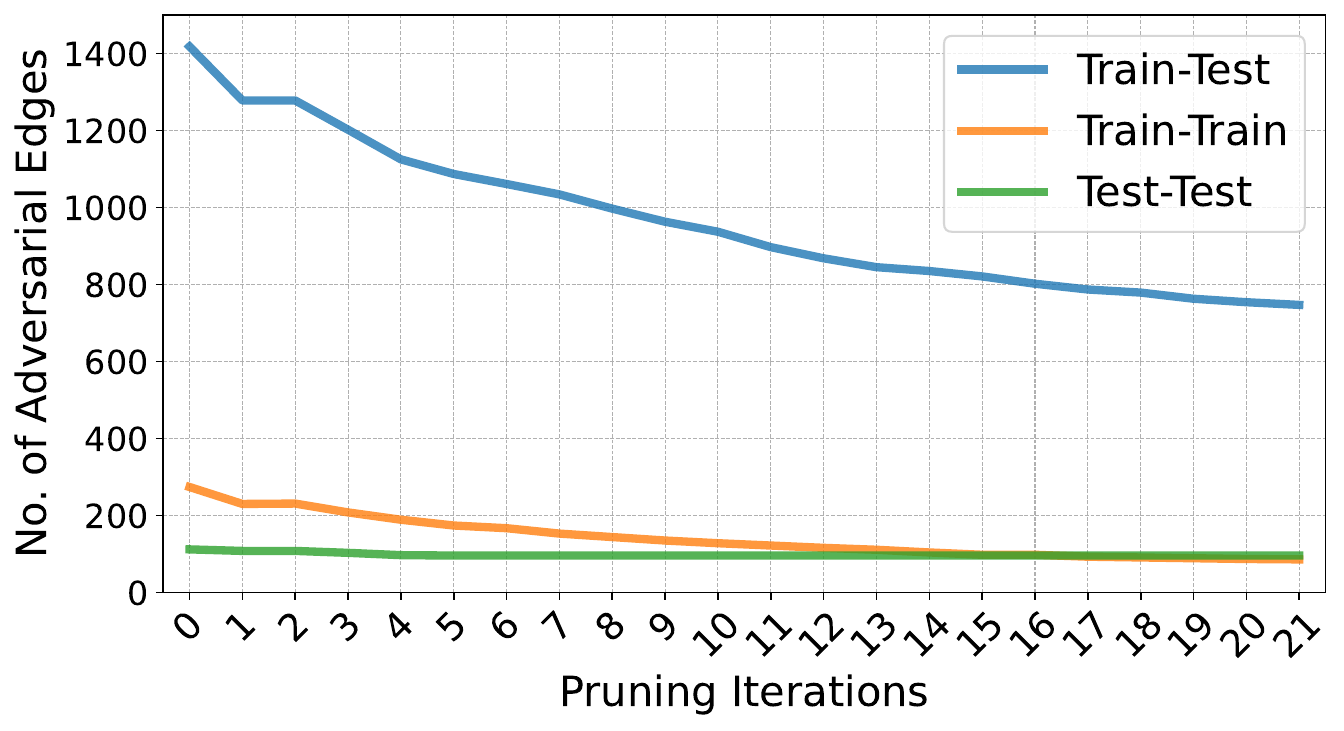}
  \end{center}
    \vspace{-10pt}
  \caption{Evolution of adversarial edges in Cora dataset (attacked by PGD, $20\%$ perturbation) as we apply ARGS to prune the graph. Train-Train edges connect two nodes from the train set. Train-Test edges connect two nodes from the train and test set, respectively. Test-Test edges connect two nodes from the test set.}
  \label{fig:edge_distribtuion}
  \vspace{-4mm}
\end{wrapfigure}

We apply these steps iteratively until we reach the desired sparsity $s_g$ and $s_{\theta}$. Algorithm~\ref{alg:rglt} illustrates our iterative pruning process yielding in ARGLTs. $||\cdot||_{0}$ is the $L_{0}$ norm counting the number of non-zero elements. As shown in Fig.~\ref{fig:edge_distribtuion} for the Cora dataset attacked by PGD attack with $20\%$ perturbation, (a) most of the adversarial perturbation edges are between train and test nodes~\citep{lirevisiting}, and (b) our proposed sparsification technique successfully removes many of the adversarial edges. In particular, after applying our technique for 20 iterations, where each iteration removes $5\%$ of the graph edges, the number of train-train, train-test, and test-test adversarial edges reduces by $68.13\%, 47.3\%$, and $14.3\%$, respectively.       

\begin{algorithm}[t]
	\caption{Adversarially Robust Graph Sparsification}
	\label{alg:rglt}
    \textbf{Input}: Graph $\mathcal{G} = \{\vect{A,X}\}$, GNN $f(\mathcal{G}, \bm{\Theta}_{0})$ with initialization $\bm{\Theta}^{0}$ , Sparsity
    \indent levels $s_g$ for graph and $s_{\theta}$ for GNN, Initial masks $\vect{m}_{g} = \vect{A}$, $\vect{m}_{\theta} = 1 \in \mathbb{R}^{||\bm{\Theta}_{0}||_{0}}$ 
%    \pierluigi{Sometimes there is zero as a subscript and sometimes I see it as a superscript. Sometimes it is in brackets. Please revise and make sure everything is correct.} 
    \\
    \textbf{Output:} Final masks $\vect{m}_{g}$,  $\vect{m}_{\theta}$
    \begin{algorithmic}[1]
    \While{ $\left(1 - \frac{||\vect{m}_{g}||_{0}}{||\vect{A}||_{0}} < s_{g}\right)$ and $ \left(1 - \frac{||\vect{m}_{\theta}||_{0}}{||\bm{\Theta}||_{0}} < s_{\theta} \right)$}
    \State $\vect{m}_{g}^{0} = \vect{m}_{g}$, $\vect{m}_{\theta}^{0} = \vect{m}_{\theta}$, $\bm{\Theta}^{0} = \{  \vect{W}^{0}_{0},  \vect{W}^{0}_{1} \} $
    \For {$t = 0, 1, 2, \ldots, T-1 $}
    \State Forward $ f(\cdot, \vect{m}_{\theta}^{t}\odot \bm{\Theta}^{t}) $ with $\mathcal{G} = \{\vect{m}_{g}^{t} \odot \vect{A}, \vect{X}\}$ to compute the loss $ \mathcal{L}_{ARGS} $ in Equation~\ref{eq:6}
    \State $\bm{\Theta}^{t+1} \leftarrow \bm{\Theta}^{t} - \mu\nabla_{\bm{\Theta}^{t}}\mathcal{L}_{ARGS} $ 
    \State $\vect{m}^{t+1}_{g} \leftarrow \vect{m}^{t}_{g} - \omega_{g}\nabla_{\vect{m}^{t}_{g}}\mathcal{L}_{ARGS} $ 
    \State  $\vect{m}^{t+1}_{\theta} \leftarrow \vect{m}^{t}_{\theta} - \omega_{\theta}\nabla_{\vect{m}^{t}_{\theta}}\mathcal{L}_{ARGS} $ 
    \EndFor
    \State $ \vect{m}_{g} = \vect{m}_{g}^{T-1} $, $ \vect{m}_{\theta
    } = \vect{m}_{\theta}^{T-1} $
    \State Set percentage $p_{g}$ of the lowest-scored values in $\vect{m}_{g}$ to 0 and set others to 1
    \State Set percentage $p_{\theta}$ of the lowest-scored values in $\vect{m}_{\theta}$ to 0 and set others to 1
    \EndWhile
    \end{algorithmic}
\end{algorithm}
% \vspace{-1mm}

\textit{Training Sparse ARGLTs.} Structure poisoning attacks do not modify the labels of the nodes and the locality structure of the test nodes is less contaminated~\citep{lirevisiting}, implying that the train node labels and the local structure of the test nodes contain relatively ``clean'' information. We leverage this insight and train the GNN sub-network using both train nodes and test nodes. We use a CE loss term for both the train $(\mathcal{L}_{0})$ and test $(\mathcal{L}_{1})$ nodes. Since the true labels of the test nodes are not available, we train an MLP on the train nodes and then use it to predict the labels for the test nodes~\citep{li2018deeper,lirevisiting}. To compute the CE loss, we use only those test nodes for which the MLP has high prediction confidence.
%We use self-learning to get the pseudo labels of the test nodes~\citep{li2018deeper,lirevisiting}. To train the GNN while reducing label noise, we use only those test nodes for which the MLP has high prediction confidence. 
The loss function used for training the sparse GNN on the sparse adjacency matrix generated by ARGS is
\begin{equation}
        \min_{\bm{\Theta}} \; \; \; \eta \mathcal{L}_{0}(f(\{\vect{m}_{g}\odot\vect{A}^{'}, \vect{X}\}, \vect{m}_{\theta}\odot\bm{\Theta})) + \zeta \mathcal{L}_{1}(f(\{\vect{m}_{g}\odot\vect{A}^{'}, \vect{X}\}, \vect{m}_{\theta}\odot\bm{\Theta}))
    \label{eq:7}
\end{equation}
where $\eta, \zeta$ are hyperparameters and $\vect{m}_{\theta}$, $\vect{m}_{g}$ are the masks evaluated by ARGS that are kept fixed throughout training. In the early pruning iterations, when graph sparsity is low, the test nodes are more useful in improving the model adversarial performance because the train nodes' localities are adversarially perturbed and there exist distribution shifts between the train and test nodes. However, as the   graph sparsity increases, adversarial edges associated with the train nodes are gradually removed by ARGS, thus reducing the distribution shift and making the contribution of the remaining train nodes more important in the adversarial training.
}

%To improve the robustness of the sparse GNN model trained on the sparse adjacency matrix, the fine-tuning optimization loss is formulated as}
% \begin{equation}
%         \min_{\bm{\Theta}} \; \; \; \eta \mathcal{L}_{0}(f(\{\vect{m}_{g}\odot\vect{A}^{'}, \vect{X}\}, \vect{m}_{\theta}\odot\bm{\Theta})) + \zeta \mathcal{L}_{1}(f(\{\vect{m}_{g}\odot\vect{A}^{'}, \vect{X}\}, \vect{m}_{\theta}\odot\bm{\Theta}))
%     \label{eq:7}
% \end{equation}
% {where $\eta, \tau$ are hyperparameters and $\vect{m}_{\theta}$, $\vect{m}_{g}$ are the masks evaluated by ARGS that are kept fixed throughout fine-tuning.}

\section{Evaluation}
\begin{figure}[!htb]
  \centering
 \includegraphics[width=1\columnwidth]{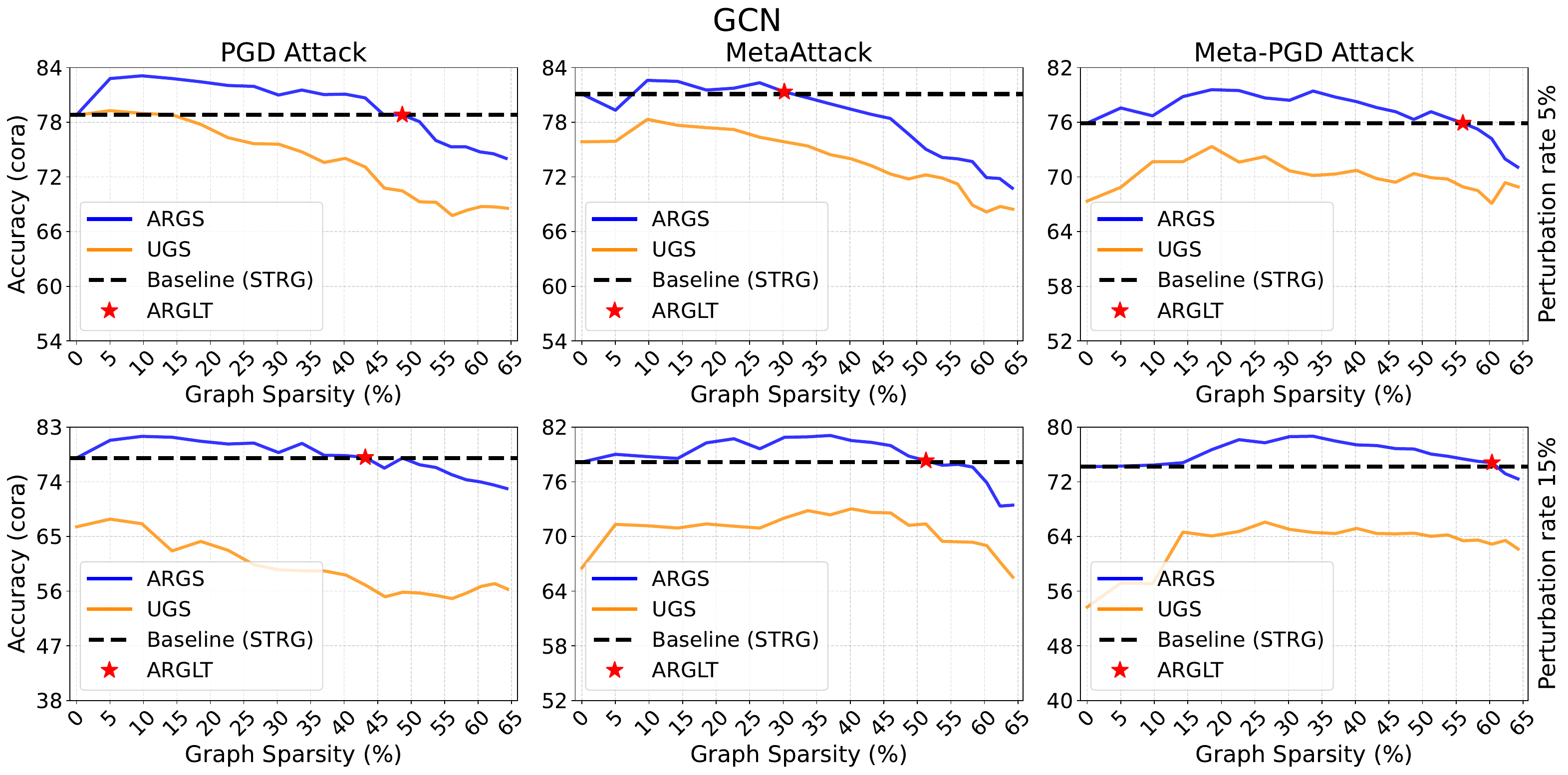}
  \includegraphics[width=1\columnwidth]{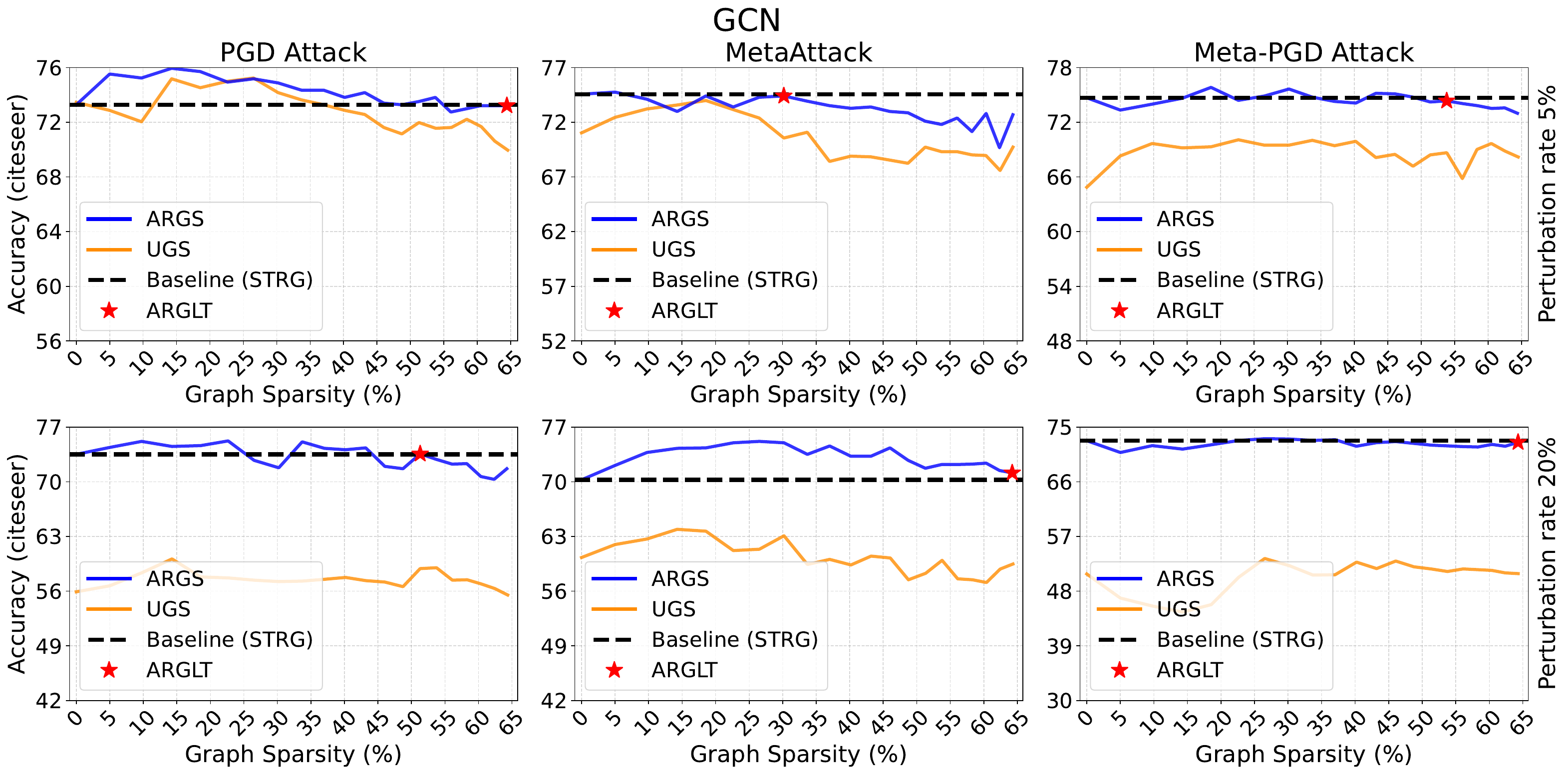}
  \includegraphics[width=1\columnwidth]{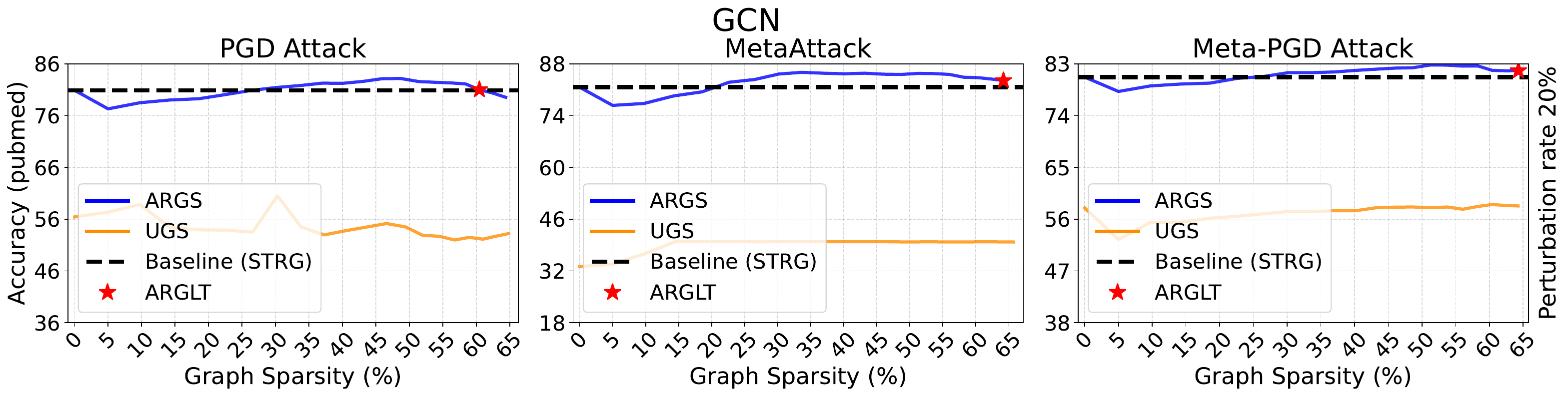}
  \caption{Node classification performance over achieved graph sparsity levels of GCNs on Cora, Citeseer, and PubMed datasets attacked by PGD, MetaAttack, and Meta-PGD with different perturbation rates. Red stars \textcolor{red}{$\star$} indicate the ARGLTs which achieve similar performance with high sparsity. Dash black lines represent the classification accuracy of the baseline method STRG~\citep{lirevisiting}.}
  \label{fig:eval_gcn}
 \end{figure}

\textbf{Evaluation Setup.} In this section, we validate the effectiveness of ARGS and the existence of adversarially robust GLTs across diverse graphs and GNN models under different adversarial attacks and perturbation rates. In particular, we evaluate our sparsification method on three widely used graph datasets, namely, Cora~\citep{mccallum2000automating}, Citeseer~\citep{sen2008collective}, and PubMed, which are attacked by three different structure poisoning attacks, namely, PGD~\citep{wu2019adversarial}, MetaAttack~\citep{zugner2020adversarial}, and Meta-PGD~\citep{mujkanovic2022defenses} with different perturbation rates, i.e., $5\%, 10\%, 15\%, 20\%$. We instead attack the large-scale graph dataset OGBN-ArXiv~\citep{hu2020open} with a more scalable attack, called h projected randomized block coordinate
descent (PR-BCD) attack~\citep{geisler2021robustness}. We compare our method with UGS~\citep{chen2021unified}, and STRG~\citep{lirevisiting}. For a fair comparison, we set $p_g = 5, p_{\theta} = 20$, similarly to the parameters used by UGS. More details on the dataset, model configurations, and hyperparameters in ARGS can be found in the appendix.

\begin{figure}[hbt!]
  \centering
 \includegraphics[width=1\columnwidth]{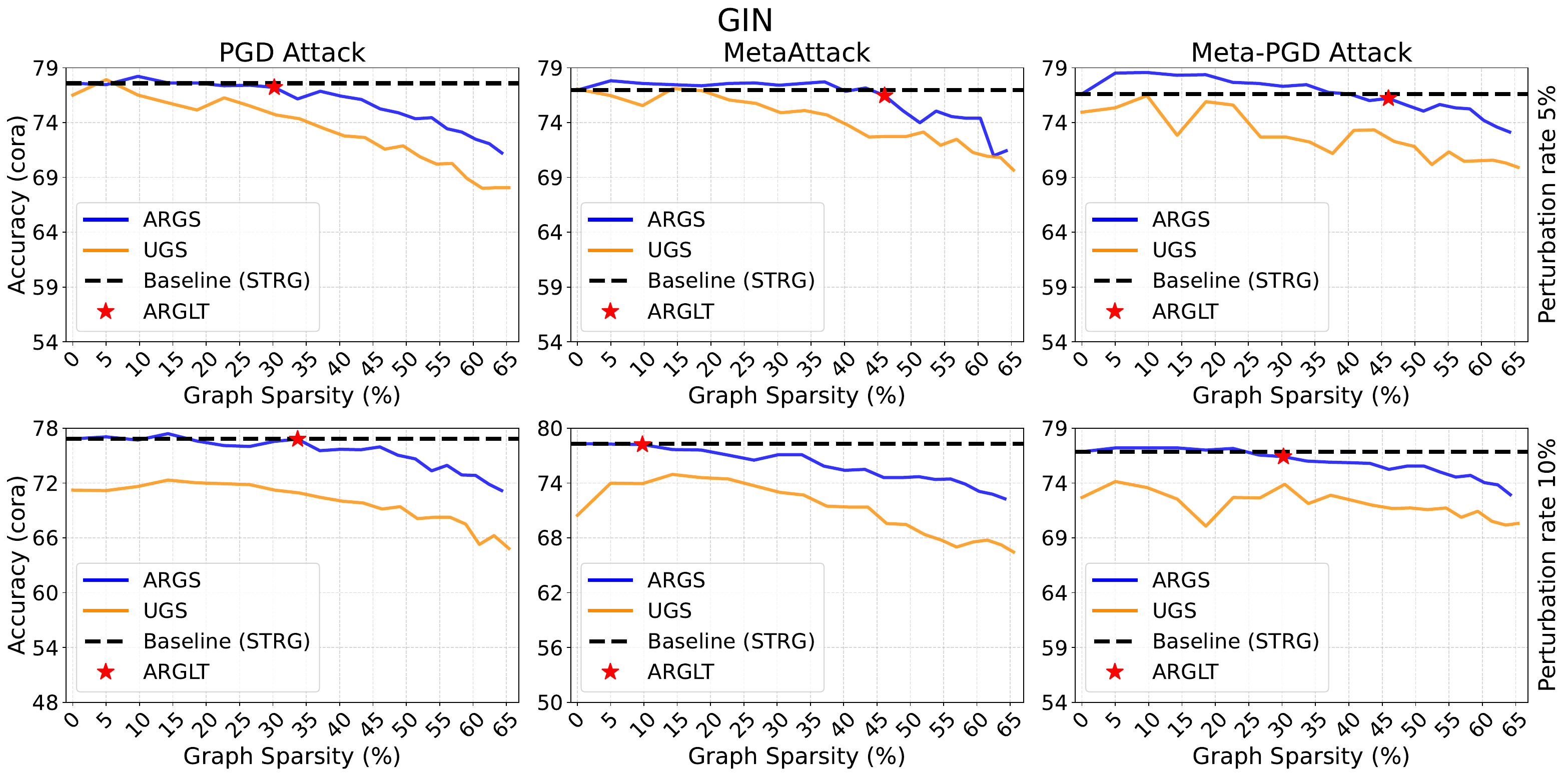}
  \includegraphics[width=1\columnwidth]{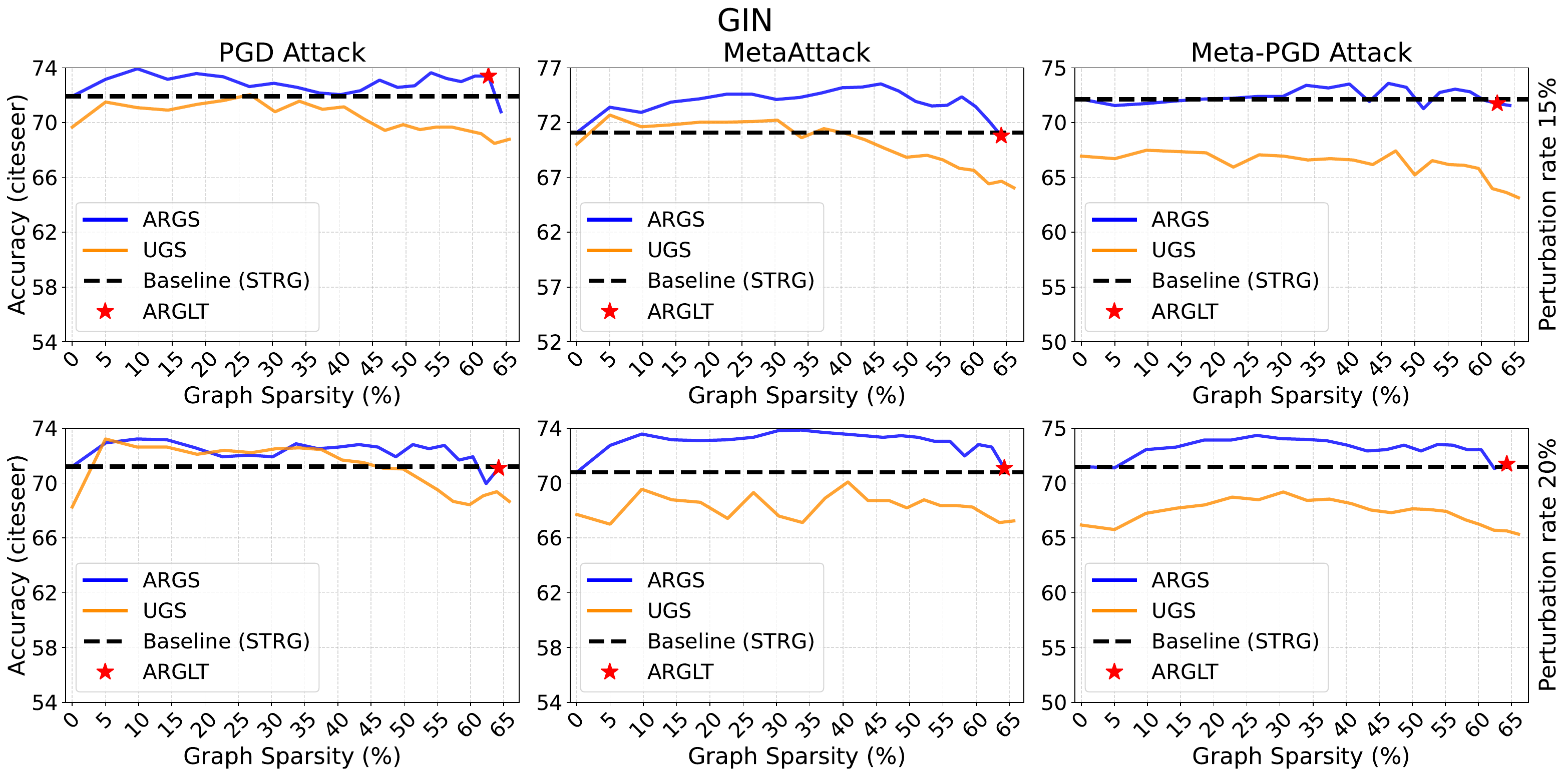}
  \caption{Node classification performance over achieved graph sparsity levels of GINs on Cora and Citeseer attacked by PGD, MetaAttack, and Meta-PGD with different perturbation rates.}
  \label{fig:eval_gin}
 \end{figure}

\setlength{\abovecaptionskip}{3pt}
\setlength{\belowcaptionskip}{3pt}
\begin{wrapfigure}{r}{0.45\textwidth}
  \begin{center}
  \vspace{-4mm}   \includegraphics[width=0.43\textwidth]{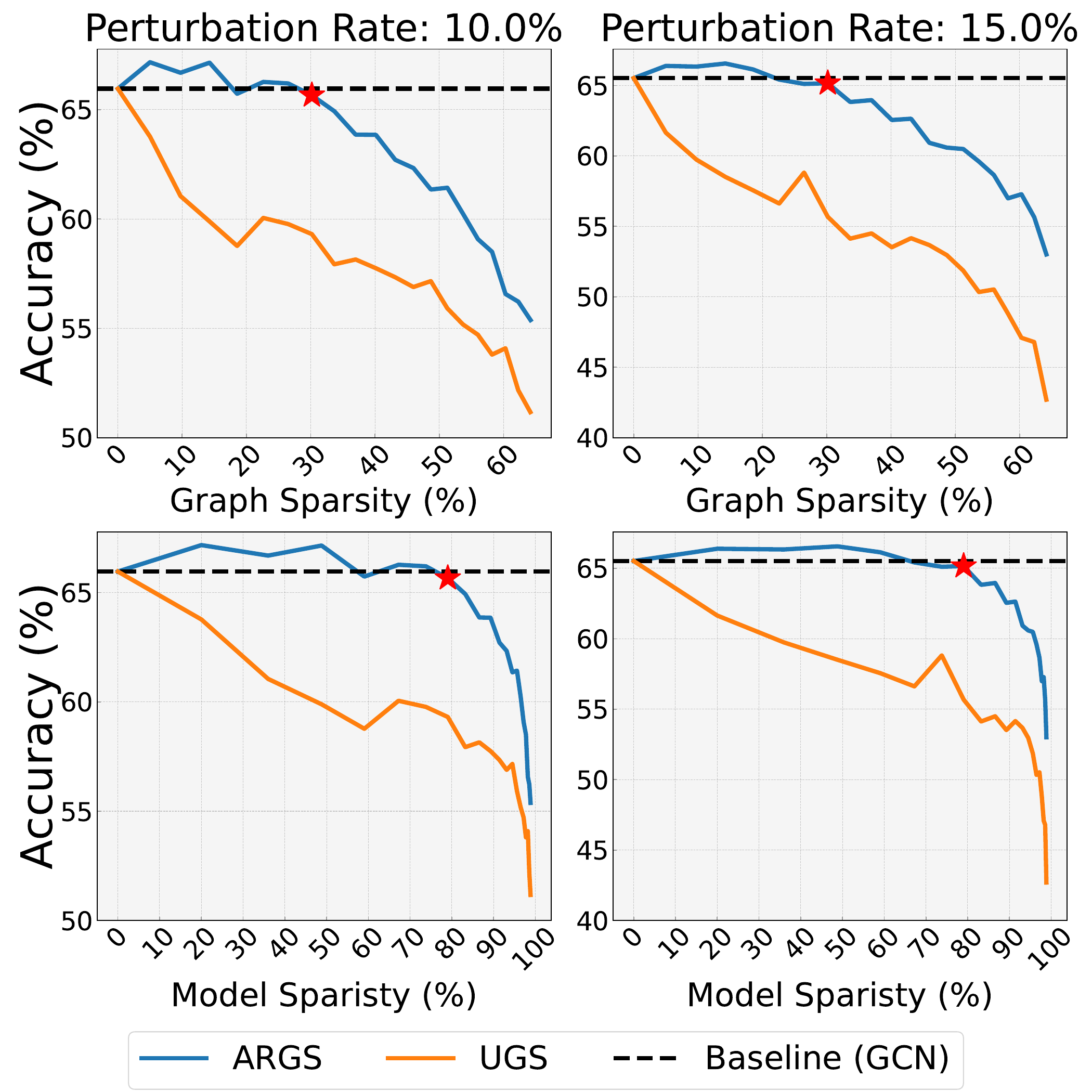}
  \end{center}
  \caption{Node classification performance versus graph sparsity and model sparsity levels for GCN on OGBN-ArXiv dataset attacked by PR-BCD.}
  \vspace{-4mm}
  \label{fig:arxiv}
\end{wrapfigure}

\textbf{Analysis of ARGS.} Figures~\ref{fig:eval_gcn} and~\ref{fig:eval_gin} compare our method with UGS in terms of the test accuracy on the 2-layer GCN and GIN models, respectively. Here we consider the test classification accuracy of STRG as the baseline. We consider different perturbation rates for the different attacks. Figure~\ref{fig:eval_gcn} shows how the test accuracy of ARGS and UGS changes with the graph sparsity and the weight sparsity during the iterative pruning process on the three datasets Cora, Citeseer, and PubMed for GCN. In each pruning iteration, the graph is pruned by $5\%$ while the GNN model weights are pruned by $20\%$. ARGLTs at a range of graph sparsity from $22.63\%$ to $60.55\%$ without performance deterioration can be identified across the GCN and GIN models. We can observe that the accuracy of our method is, on average, notably higher than the one of UGS. For example, in the case of the Cora dataset attacked by PGD with a perturbation budget of $15\%$,  when the graph sparsity is $48.7\%$ and the GNN weight sparsity is $94.61\%$, the accuracy of our technique is $77.92\%$ whereas the accuracy of UGS is $55.84\%$. Therefore, for the same sparsity level, ARGS can achieve $22.08\%$ better classification accuracy than UGS. Moreover, the higher accuracy of ARGS shows that the ARGLTs are more robust than those identified by UGS. In the case of the PubMed dataset, the accuracy of our method is stable even when the graph sparsity is high. This is due to the higher density of the PubMed graph allowing for more edges to be removed while maintaining accuracy. Similarly to GCN, ARGLTs can be identified by ARGS with high graph and model sparsity for GIN, as shown in Figure~\ref{fig:eval_gin}. 

We also evaluate the robustness of ARGS on the large-scale dataset OGBN-ArXiv. We use the PR-BCD attack for perturbing the dataset and the reference GNN model is 28-layer ResGCN~\citep{li2020deepergcn}. PGD or MetaAttack face timeout due to memory for these large graphs. Figure~\ref{fig:arxiv} shows that ARGS is able to identify ARGLTs that have high model and graph sparsity. In particular, the model sparsity and graph sparsity are $79.03\%$ and $30.17\%$ for the 10\% perturbed dataset, and $78.31\% $ and $30.81\%$, respectively, for the 15\% perturbed dataset. These results show that ARGS can find highly sparse GLTs also for large-scale graph datasets.

\textbf{Ablation Study.}
To verify the effectiveness of each component of the proposed loss function used for the sparsification algorithm,  we perform an ablation study as shown in Table~\ref{ablation study}. We consider the Cora dataset under PGD attack with 10\% and 20\% perturbation rates. Configuration 1 corresponds to ARGS with all the loss components in~\eqref{eq:6}. In configuration 2, we do not use the feature smoothness component in \eqref{eq:4} while performing the sparsification. In configuration 3, we skip the CE loss associated with the predicted test nodes in \eqref{eq:5}, and in configuration 4 we skip both the smoothness and CE loss on the predicted test nodes. As shown in Table \ref{ablation study}, both configurations 2 and 3 improve the final performance compared to that of configuration 4, highlighting the importance of the losses introduced in \eqref{eq:4} and \eqref{eq:5}. More importantly, at both high and low target sparsity, we yield the best classification performance with configuration 1, showcasing the importance of the unified loss function in \eqref{eq:6}. Further, ablation studies on different datasets and on the loss components associated with the loss function in \eqref{eq:7} are provided in the appendix. 

% abaltion study table
\begin{table}[htbp]
  \centering
  \caption{Ablation Study}
  \label{ablation study}
  \begin{adjustbox}{center}
    \tiny
    \begin{tabular}{c|c|c|c|c|c|c|c|c|c}
      \toprule
      \multicolumn{6}{c|}{GCN, Cora, PGD Attack}  
      & 
      \multicolumn{2}{c|}{\parbox{4cm}{\centering Classification Accuracy at \\ Perturbation Rate 10\%}}  
      & 
      \multicolumn{2}{c}{\parbox{4cm}{\centering Classification Accuracy at \\ Perturbation Rate 20\%}} \\
      \midrule
      Configuration & $\alpha$ & $\beta$ & $\gamma$ & $\eta$ & $\zeta$ & {\parbox{2cm}{Graph Sparsity 9.8\% \\  Model Sparsity 36.1\%}}  & {\parbox{2cm}{Graph Sparsity 64.4\% \\  Model Sparsity 98.9\%}}  & {\parbox{2cm}{Graph Sparsity 9.8\% \\  Model Sparsity 36.1\%}} & {\parbox{2cm}{Graph Sparsity 64.5\% \\  Model Sparsity 98.9\%}} \\
      \midrule
      1 & \ding{51} & \ding{51} & \ding{51} & \ding{51} & \ding{51} & \textbf{83.25} & \textbf{75.10} & \textbf{80.63} & \textbf{75.60} \\
      2 & \ding{51} & \ding{55} & \ding{51} & \ding{51} & \ding{51} & 82.04 & 70.57 & 78.92 & 64.84 \\
      3 & \ding{51} & \ding{51} & \ding{55} & \ding{51} & \ding{51} & 82.44 & 72.84 & 78.97 & 52.92 \\
      4 & \ding{51} & \ding{55} & \ding{55} & \ding{51} & \ding{51} & 80.58 & 62.42 & 75.7 & 54.18 \\
      %5 & \ding{51} & \ding{51} & \ding{51} & \ding{51} & \ding{55} & 79.63 & 71.83 & 66.65 & 58.4 \\
      %6 & \ding{51} & \ding{51} & \ding{55} & \ding{51} & \ding{55} & 73.79 & 68.36 & 63.98 & 45.17 \\
    \bottomrule
    \end{tabular}
  \end{adjustbox}
\end{table}

\section{Conclusion}
In this paper, we first empirically observed that the performance of GLTs collapses against structure perturbation poisoning attacks. To address this issue, we presented a new adversarially robust graph sparsification technique, ARGS, that prunes the perturbed adjacency matrix and the GNN weights by optimizing a novel loss function.  By iteratively applying ARGS, we found ARGLTs that are highly sparse yet achieve competitive performance under different structure poisoning attacks. Our evaluation showed the effectiveness of our
method over UGS at both high and low-sparsity regimes. Future work involves the exploration of adversarially robust graph sparsification techniques in the case of graphs where the homophily property does not hold.

% \section{Broader Impact}
% Graphs are universal structures of real-world complex systems. Empowering deep learning for reasoning and predicting over graph-structured data is of broad interest and wide applications, such as recommendation systems, neural architecture search, and drug discovery. However, scaling up GNNs to large datasets is often impacted due to high computational footprints. Besides, recent studies found that GNNs are highly vulnerable to adversarial attacks. This work is in an effort to find GNN models that reduce the computational footprints yet maintain adversarial robustness. Finding such robust models will enable efficient and reliable deployment of GNNs.

\section*{Acknowledgments} 
    This research was supported in part by the National Science Foundation under Awards 1846524 and 2139982, the Office of Naval Research under Award N00014-20-1-2258, the Okawa Research Grant, and the USC Center for Autonomy and Artificial Intelligence. 

\bibliography{references}{}
\bibliographystyle{plain}

\section{Appendix}
In Section~\ref{dataset}, we provide more details about the datasets used in the main sections of the paper. We describe the model and attack settings in Section~\ref{implementation}. More experiments and ablation studies 
% that are not included in the main paper due to space constraints 
are presented in Section~\ref{more_results}. 

\subsection{Dataset Details}\label{dataset}

We use three commonly used benchmark datasets, namely, Cora, Citeseer, PubMed, together with the OGBN-ArXiv to evaluate the efficacy of ARGS in finding ARGLTs. The details are summarized in Table~\ref{dataset_stat}. We use the data split $10\%/10\%/80\%$ (Train/Validation/Test) for these datasets and consider the largest connected component (LCC).
For OGBN-ArXiv we follow the data split setting of Open Graph Benchmark (OGB)~\citep{hu2020open}.

\begin{table}[htbp]
\centering
\caption{Dataset details}
\label{dataset_stat}
\begin{tabular}{c | c c c c}
 \hline\hline
 Datasets & \#Nodes  & \#Edges & Classes & Features \\ 
 \hline\hline
 Cora & 2485 & 5069 & 7 & 1433 \\ 
 Citeseer & 2110 & 3668 & 6 & 3703 \\
 PubMed & 19717 & 44338 & 3 & 500 \\
  OGBN-ArXiv  & 169343 & 1166243 & 40 & 128 \\
 \hline
\end{tabular}
\end{table}

%Different adversarial attacks like PGD, MetaAttack, and PGD-Meta attack face OOM issues for large-scale datasets from the Open Graph Benchmark (OGB).

\subsection{Implementation Details}\label{implementation}

For a fair comparison, we follow the setup used by UGS as our default setting. For Cora, Citeseer, and PubMed we conduct all our experiments on two-layer GCN/GIN networks with 512 hidden units. The graph sparsity $p_g$ and model sparsity $p_{\theta}$ are $5\%$ and $20\%$ unless otherwise stated. The value of $\beta$ is chosen from $[0, 0.01, 0.1, 1]$ while the value of $\alpha, \gamma, \eta, $ and $\zeta$  is $1$ by default. We use the Adam optimizer for training the GNNs. The value of $\lambda_1$ and $\lambda_2$ is $10^{-2}$, $10^{-2}$ for Cora and Citeseer, while for PubMed it is 1e-6, 1e-3, respectively. In each pruning round, the number of epochs to update the masks is by default 200 and we use early stopping. The 2-layer MLP used for predicting the pseudo labels of the test nodes has a hidden dimension of 1024. We use DeepRobust, an adversarial attack repository~\citep{li2020deeprobust}, to implement the PGD attack and MetaAttack on the Cora, Citeseer, and PubMed dataset for perturbation rates $5\%, 10\%, 15\%$ and $20\%$. In case of the PGD-Meta attack, we use the code provided by the authors to perform the attack~\citep{mujkanovic2022defenses}. We face OOM issues when applying these attacks on large-scale graph datasets. We use Pytorch-Geometric~\citep{hu2020open} for performing the PR-BCD attack on the OGBN-ArXiv dataset. An NVIDIA Tesla V100 32GB GPU is used to conduct all our experiments.

\subsection{Ablation Study}\label{ablation}

We perform an ablation study to verify the effectiveness of each component of the proposed loss function used for the sparsification algorithm. A part of this ablation study is shown in Table~\ref{ablation study}. In this section, we present the evaluations performed on the Cora dataset for all the 3 different attacks with all the 4 different perturbation rates. Configuration 1 corresponds to ARGS executed with all the loss components. As shown in Table \ref{ablation_study_sup} and \ref{ablation_study_sup_2}, at both high and low target sparsity, we yield the best classification performance with configuration 1, underscoring the importance of the selected loss function. 

%Further, ablation studies on different datasets and on the loss components associated with the loss function of Eq.~\eqref{eq:7} are provided in the supplementary material.

%%%%%%%%%%%%%%%%%%%%%%%%%%%%%%%%%%%%%%%%%%%%%%%%%%%%%%%%%%%%
% GCN ablation study table
\begin{table}[h]
  \centering
  \caption{Ablation Study}
  \label{ablation_study_sup}
    \begin{adjustbox}{center}
    \tiny
    \begin{tabular}{c|c|c|c|c|c|c|c|c|c}
      \toprule
      \multicolumn{6}{c|}{GCN, Cora, PGD Attack}  
      & 
      \multicolumn{2}{c|}{\parbox{4cm}{\centering Classification Accuracy at \\ Perturbation Rate 5\%}}  
      & 
      \multicolumn{2}{c}{\parbox{4cm}{\centering Classification Accuracy at \\ Perturbation Rate 15\%}} \\
      \midrule
      Configuration & $\alpha$ & $\beta$ & $\gamma$ & $\eta$ & $\zeta$ & {\parbox{2cm}{Graph Sparsity 22.7\% \\  Model Sparsity 67.7\%}}  & {\parbox{2cm}{Graph Sparsity 60.4\% \\  Model Sparsity 98.2\%}}  & {\parbox{2cm}{Graph Sparsity 22.7\% \\  Model Sparsity 67.7\%}} & {\parbox{2cm}{Graph Sparsity 60.4\% \\  Model Sparsity 98.2\%}} \\
      \midrule
        1 & \ding{51} & \ding{51} & \ding{51} & \ding{51} & \ding{51} & \textbf{82.04} & \textbf{74.75} & \textbf{80.23} & \textbf{73.99} \\
        2 & \ding{51} & \ding{55} & \ding{51} & \ding{51} & \ding{51} & 81.84 & 73.64 & 79.98 & 68.81 \\
        3 & \ding{51} & \ding{51} & \ding{55} & \ding{51} & \ding{51} & 81.69 & 74.45 & 76.86 & 72.89 \\
        4 & \ding{51} & \ding{55} & \ding{55} & \ding{51} & \ding{51} & 79.28 & 71.33 & 74.70 & 63.48 \\
    \bottomrule
    \end{tabular}
    \end{adjustbox}

    \begin{adjustbox}{center}
    \tiny
    \begin{tabular}{c|c|c|c|c|c|c|c|c|c}
      \toprule
      \multicolumn{6}{c|}{GCN, Cora, MetaAttack}  
      & 
      \multicolumn{2}{c|}{\parbox{4cm}{\centering Classification Accuracy at \\ Perturbation Rate 5\%}}  
      & 
      \multicolumn{2}{c}{\parbox{4cm}{\centering Classification Accuracy at \\ Perturbation Rate 10\%}} \\
      \midrule
      Configuration & $\alpha$ & $\beta$ & $\gamma$ & $\eta$ & $\zeta$ & 
        {\parbox{2cm}{Graph Sparsity 22.7\% \\  Model Sparsity 67.6\%}}  & {\parbox{2cm}{Graph Sparsity 62.3\% \\  Model Sparsity 98.6\%}}  & {\parbox{2cm}{Graph Sparsity 22.6\% \\  Model Sparsity 67.5\%}} & {\parbox{2cm}{Graph Sparsity 64.2\% \\  Model Sparsity 98.9\%}} \\
      \midrule
        1 & \ding{51} & \ding{51} & \ding{51} & \ding{51} & \ding{51} & \textbf{81.74} & \textbf{71.83} & \textbf{80.23} & \textbf{71.58} \\
        2 & \ding{51} & \ding{55} & \ding{51} & \ding{51} & \ding{51} & 80.89 & 69.91 & 78.17 & 70.98 \\
        3 & \ding{51} & \ding{51} & \ding{55} & \ding{51} & \ding{51} & 79.88 & 71.33 & 75.40 & 66.81 \\
        4 & \ding{51} & \ding{55} & \ding{55} & \ding{51} & \ding{51} & 78.89 & 69.03 & 75.40 & 60.97 \\
    \bottomrule
    \end{tabular}
    \end{adjustbox}
\end{table}

\begin{table}[h]
  \centering
  \caption{Ablation Study}
  \label{ablation_study_sup_2}
        \begin{adjustbox}{center}
    \tiny
    \begin{tabular}{c|c|c|c|c|c|c|c|c|c}
      \toprule
      \multicolumn{6}{c|}{GCN, Cora, MetaAttack}  
      & 
      \multicolumn{2}{c|}{\parbox{4cm}{\centering Classification Accuracy at \\ Perturbation Rate 15\%}}  
      & 
      \multicolumn{2}{c}{\parbox{4cm}{\centering Classification Accuracy at \\ Perturbation Rate 20\%}} \\
      \midrule
      Configuration & $\alpha$ & $\beta$ & $\gamma$ & $\eta$ & $\zeta$ & 
      {\parbox{2cm}{Graph Sparsity 22.7\% \\  Model Sparsity 67.6\%}}  & {\parbox{2cm}{Graph Sparsity 60.4\% \\  Model Sparsity 98.2\%}}  & {\parbox{2cm}{Graph Sparsity 22.6\% \\  Model Sparsity 67.5\%}} & {\parbox{2cm}{Graph Sparsity 60.3\% \\  Model Sparsity 98.2\%}} \\
      \midrule
        1 & \ding{51} & \ding{51} & \ding{51} & \ding{51} & \ding{51} & \textbf{80.73} & \textbf{75.91} & \textbf{79.38} & \textbf{70.37} \\
        2 & \ding{51} & \ding{55} & \ding{51} & \ding{51} & \ding{51} & 80.23 & 73.69 & 78.72 & 69.97 \\
        3 & \ding{51} & \ding{51} & \ding{55} & \ding{51} & \ding{51} & 77.97 & 72.74 & 75.50 & 69.16 \\
        4 & \ding{51} & \ding{55} & \ding{55} & \ding{51} & \ding{51} & 78.42 & 72.08 & 74.09 & 68.86 \\
    \bottomrule
    \end{tabular}
    \end{adjustbox}
    \begin{adjustbox}{center}
    \tiny
    \begin{tabular}{c|c|c|c|c|c|c|c|c|c}
      \toprule
      \multicolumn{6}{c|}{GCN, Cora, Meta-PGD Attack}  
      & 
      \multicolumn{2}{c|}{\parbox{4cm}{\centering Classification Accuracy at \\ Perturbation Rate 5\%}}  
      & 
      \multicolumn{2}{c}{\parbox{4cm}{\centering Classification Accuracy at \\ Perturbation Rate 10\%}} \\
      \midrule
      Configuration & $\alpha$ & $\beta$ & $\gamma$ & $\eta$ & $\zeta$ & {\parbox{2cm}{Graph Sparsity 22.7\% \\  Model Sparsity 67.7\%}}  & {\parbox{2cm}{Graph Sparsity 60.4\% \\  Model Sparsity 98.2\%}}  & {\parbox{2cm}{Graph Sparsity 22.7\% \\  Model Sparsity 67.6\%}} & {\parbox{2cm}{Graph Sparsity 60.3\% \\  Model Sparsity 98.2\%}} \\
      \midrule
        1 & \ding{51} & \ding{51} & \ding{51} & \ding{51} & \ding{51} & \textbf{79.48} & \textbf{74.20} & \textbf{77.01} & \textbf{73.69} \\
        2 & \ding{51} & \ding{55} & \ding{51} & \ding{51} & \ding{51} & 78.12 & 73.99 & 76.87 & 70.67 \\
        3 & \ding{51} & \ding{51} & \ding{55} & \ding{51} & \ding{51} & 76.86 & 69.67 & 76.01 & 69.97 \\
        4 & \ding{51} & \ding{55} & \ding{55} & \ding{51} & \ding{51} & 76.46 & 68.67 & 75.51 & 68.46 \\
    \bottomrule
    \end{tabular}
    \end{adjustbox}
    \begin{adjustbox}{center}
    \tiny
    \begin{tabular}{c|c|c|c|c|c|c|c|c|c}
      \toprule
      \multicolumn{6}{c|}{GCN, Cora, Meta-PGD Attack}  
      & 
      \multicolumn{2}{c|}{\parbox{4cm}{\centering Classification Accuracy at \\ Perturbation Rate 15\%}}  
      & 
      \multicolumn{2}{c}{\parbox{4cm}{\centering Classification Accuracy at \\ Perturbation Rate 20\%}} \\
      \midrule
      Configuration & $\alpha$ & $\beta$ & $\gamma$ & $\eta$ & $\zeta$ & {\parbox{2cm}{Graph Sparsity 22.7\% \\  Model Sparsity 67.4\%}}  & {\parbox{2cm}{Graph Sparsity 60.4\% \\  Model Sparsity 98.2\%}}  & {\parbox{2cm}{Graph Sparsity 22.7\% \\  Model Sparsity 67.4\%}} & {\parbox{2cm}{Graph Sparsity 60.4\% \\  Model Sparsity 98.2\%}} \\
      \midrule
        1 & \ding{51} & \ding{51} & \ding{51} & \ding{51} & \ding{51} & \textbf{78.17} & \textbf{74.80} & \textbf{78.22} & \textbf{75.55} \\
        2 & \ding{51} & \ding{55} & \ding{51} & \ding{51} & \ding{51} & 77.31 & 73.24 & 76.70 & 73.59 \\
        3 & \ding{51} & \ding{51} & \ding{55} & \ding{51} & \ding{51} & 77.26 & 70.88 & 77.46 & 67.25 \\
        4 & \ding{51} & \ding{55} & \ding{55} & \ding{51} & \ding{51} & 75.80 & 69.92 & 75.92 & 66.80 \\
    \bottomrule
    \end{tabular}
    \end{adjustbox}
    % \begin{adjustbox}{center}
    %   \tiny
    %   \begin{tabular}{c|c|c|c|c|c|c|c|c|c}
    %     \toprule
    %     \multicolumn{6}{c|}{GCN, Citeseer, PGD Attack}  
    %     & 
    %     \multicolumn{2}{c|}{\parbox{4cm}{Classification Accuracy at \\ Perturbation Rate 10\%}}  
    %     & 
    %     \multicolumn{2}{c}{\parbox{4cm}{Classification Accuracy at \\ Perturbation Rate 15\%}} \\
    %     \midrule
    %     Configuration & $\alpha$ & $\beta$ & $\gamma$ & $\eta$ & $\zeta$ & {Graph Sparsity 9.8\%}  & {Graph Sparsity 64.3\%} & {Graph Sparsity 9.8\%} & {Graph Sparsity 64.2\%} \\
    %     \midrule
    %     1 & \ding{51} & \ding{51} & \ding{51} & \ding{51} & \ding{51} & \textbf{75.89} & 71.68 & \textbf{75.24} & \textbf{71.98} \\
    %     2 & \ding{51} & \ding{55} & \ding{51} & \ding{51} & \ding{51} & 75.41 & 69.67 & 74.76 & 70.62 \\
    %     3 & \ding{51} & \ding{51} & \ding{55} & \ding{51} & \ding{51} & 74.29 & 68.96 & 73.1 & 65.64 \\
    %     4 & \ding{51} & \ding{55} & \ding{55} & \ding{51} & \ding{51} & 72.39 & 69.37 & 74.05 & 67.89 \\
    %     % Configuration1 & \ding{51} & \ding{51} & \ding{51} & \ding{51} & \ding{55} & 72.04 & \textbf{71.86} & 71.09 & 71.03 \\
    %     % Configuration1 & \ding{51} & \ding{51} & \ding{55} & \ding{51} & \ding{55} & 70.44 & 67.24 & 67.65 & 63.92 \\
    %     \bottomrule
    %   \end{tabular}
    % \end{adjustbox}
\end{table}

\subsection{Additional Results}\label{more_results}

\begin{figure}
  \centering
  \label{fig:res}
 \includegraphics[width=1\columnwidth]{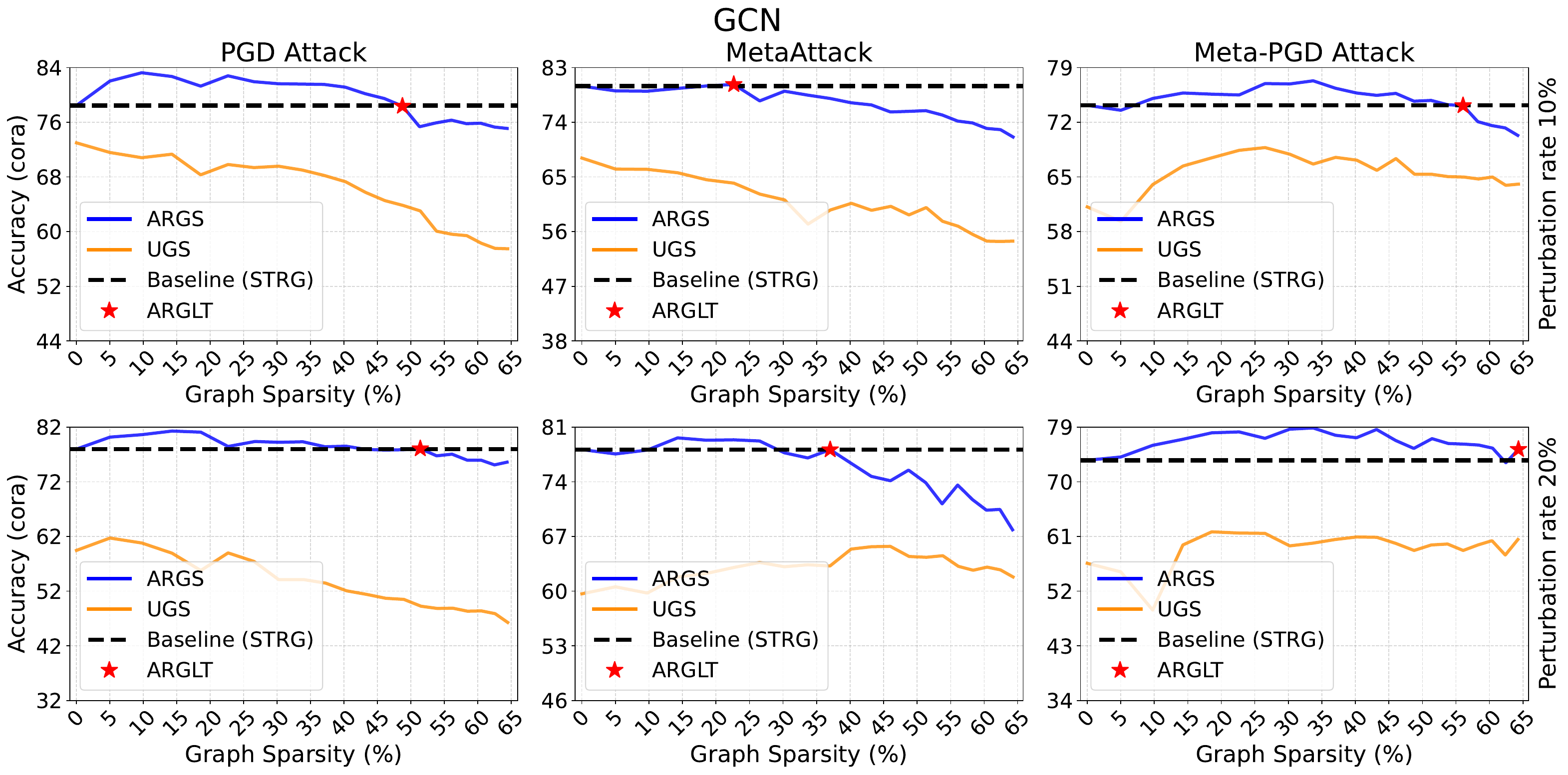}
 \includegraphics[width=1\columnwidth]{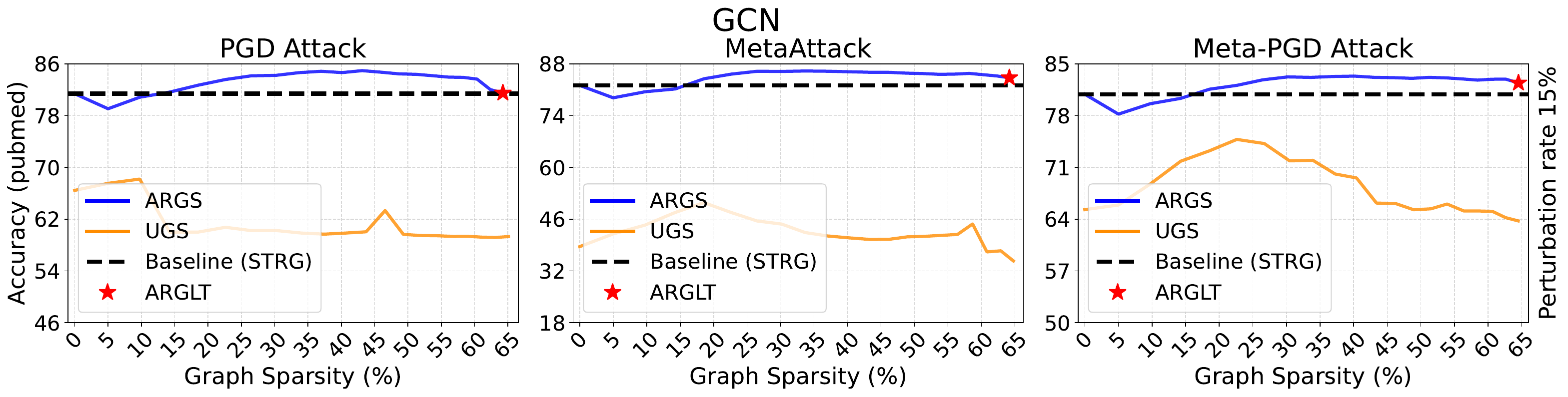}
  \caption{Node classification performance over achieved graph sparsity levels of GCNs on Cora and PubMed attacked by PGD, MetaAttack, and Meta-PGD with different perturbation rates.}
  \label{fig:eval_gcn_append}
 \end{figure}

Figure~\ref{fig:eval_gcn_append} shows the performance of ARGS in terms of the test accuracy on 2-layer GCN models. More specifically, in addition to the results with $5\%$ and $15\%$ perturbation rates with GCN for the 3 different attacks in Figure~\ref{fig:eval_gcn}, we also include the results for $10\%$ and $20\%$ perturbation rates as well as the performance for the PubMed dataset.

\end{document}